\pdfoutput=1
\def\year{2018}\relax
\documentclass[letterpaper]{article} 
\usepackage{aaai18-arxiv}  
\usepackage{times}  
\usepackage{helvet}  
\usepackage{courier}  
\usepackage{url}  
\usepackage{graphicx}  
\newcommand{\eat}[1]{}
\usepackage{amsthm}
\usepackage{amssymb}
\usepackage{amsmath}
\usepackage[ruled,vlined]{algorithm2e}
\usepackage{url}
\usepackage{booktabs} 
\newcommand\todo[1]{\textcolor{red}{#1}}
\newcommand{\topic}[1]{\noindent \underline{ \bf #1}}
\newtheorem{theorem}{Theorem}
\usepackage{xcolor}
\usepackage{enumitem}
\usepackage{subfig}

\DeclareMathOperator*{\argmax}{arg\,max}
\usepackage{hyperref}
\renewcommand{\footnotesize}{\scriptsize}

\begin{document}
\title{Feature Engineering for Predictive Modeling using Reinforcement Learning}


\author{Udayan Khurana, Horst Samulowitz, Deepak Turaga\\
\{ukhurana,samulowitz,turaga\}@us.ibm.com\\
IBM TJ Watson Research Center\\
}

\maketitle

\begin{abstract}
Feature engineering is a crucial step in the process of predictive modeling. It involves the transformation of given feature space, typically using mathematical functions, with the objective of reducing the modeling error for a given target. 
However, there is no well-defined basis for performing effective feature engineering. It involves domain knowledge, intuition, and most of all, a lengthy process of trial and error.
The human attention involved in overseeing this process significantly influences the cost of model generation.
We present a new framework to automate feature engineering. 
It is based on performance driven exploration of a {\em transformation graph}, which systematically and compactly enumerates the space of given options. A highly efficient exploration strategy is derived through reinforcement learning on past examples.
\end{abstract}

\section{Introduction}

Predictive analytics are widely used in support for decision making across a variety of domains including fraud detection, marketing, drug discovery, advertising, risk management, amongst several others. Predictive models are constructed using supervised learning algorithms 
where classification or regression models are trained on historical data to predict future outcomes. 
The underlying representation of the data is crucial for the learning algorithm to work effectively.
In most cases, appropriate transformation of data is an essential prerequisite step before model construction. 

\begin{figure}
\begin{center}
\subfloat[Original data]{\includegraphics[width=120pt, height=120pt]{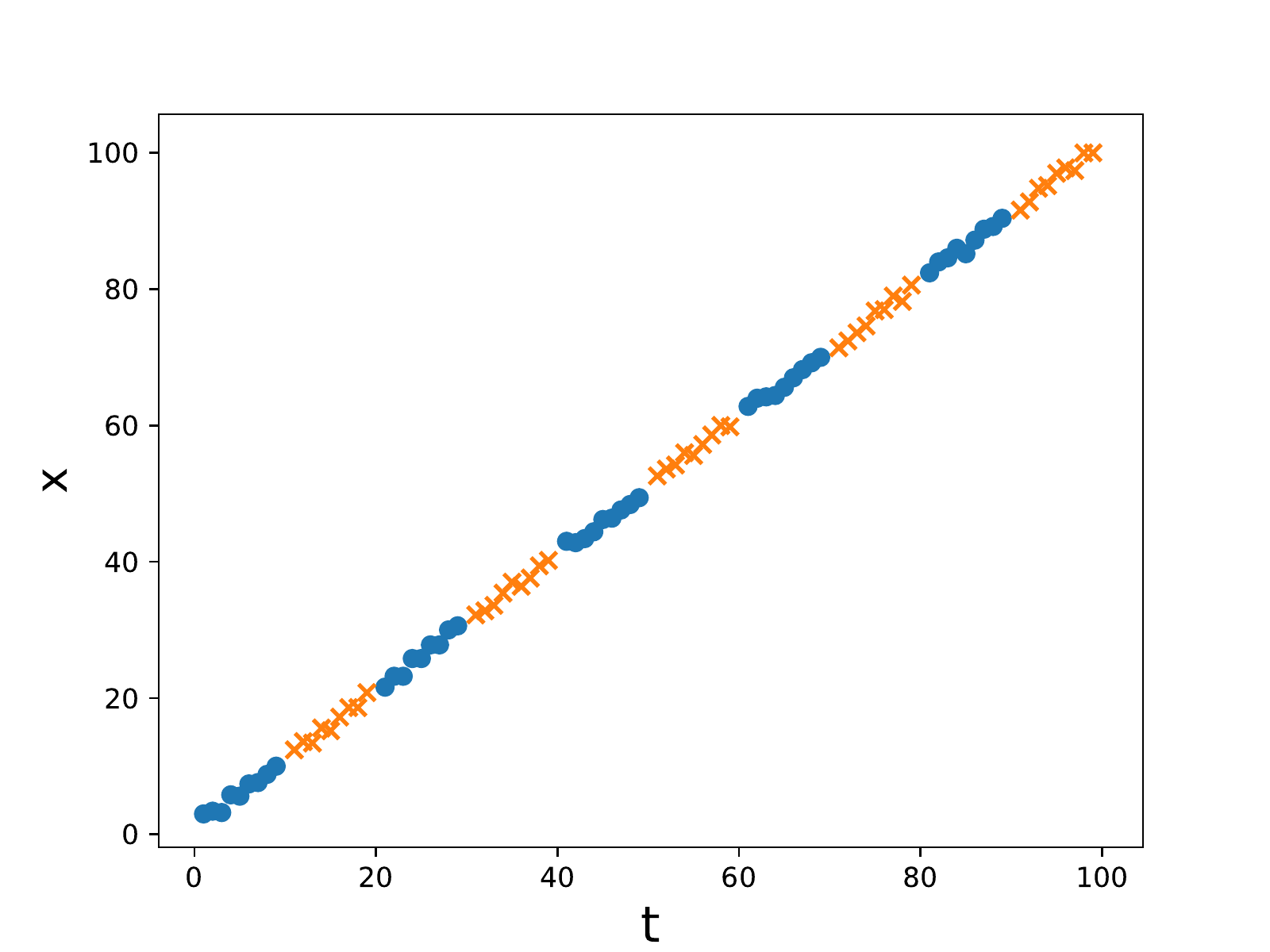}}
\subfloat[Engineered data.]{\includegraphics[width=120pt, height=120pt]{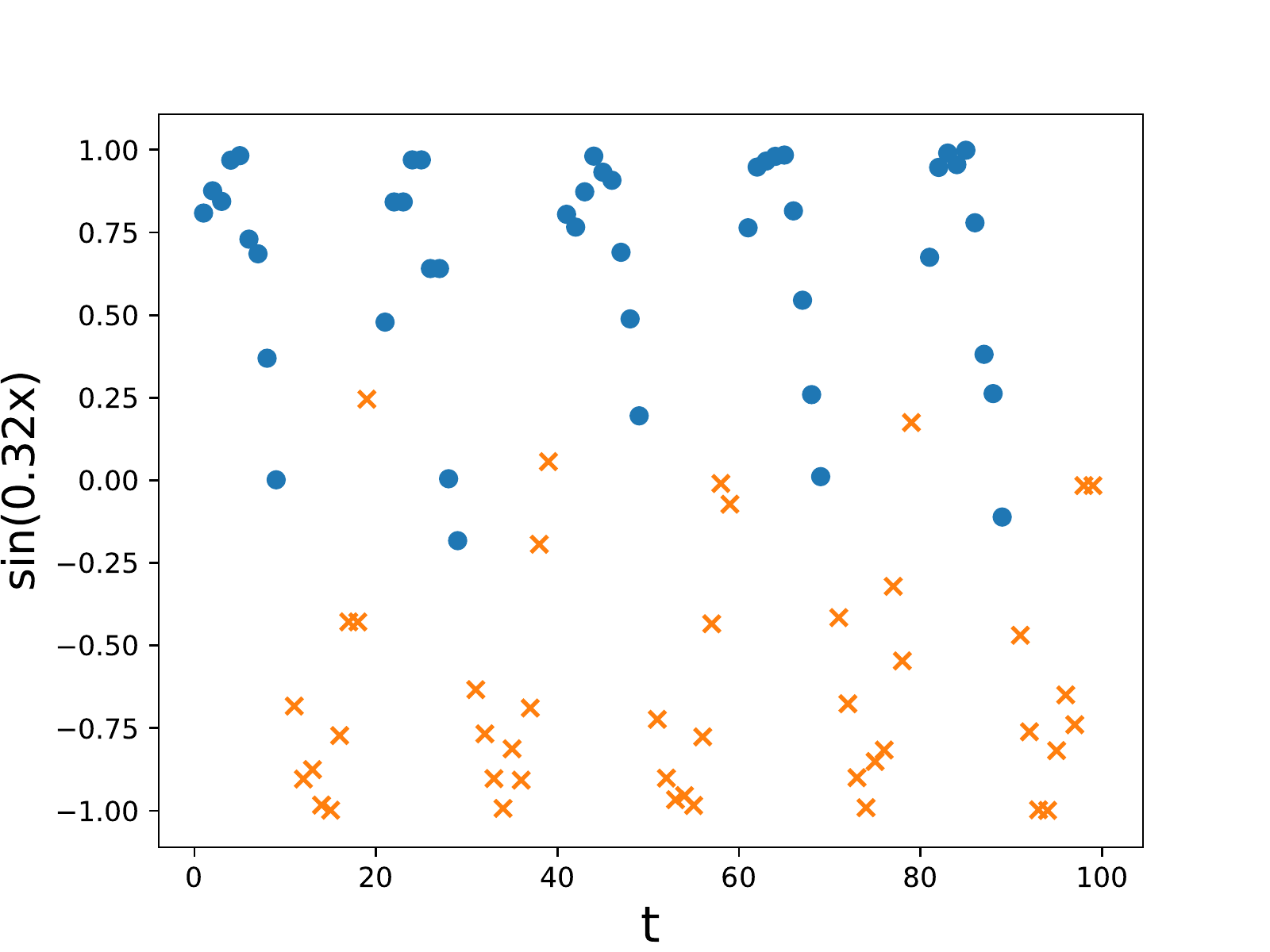}}
\end{center}
\caption[Illustration of different representation choices.]{Illustration of different representation choices.}
\label{fig:ex1}
\end{figure}

\begin{figure}
\subfloat[Original features and target count.]{\includegraphics[width=\linewidth]{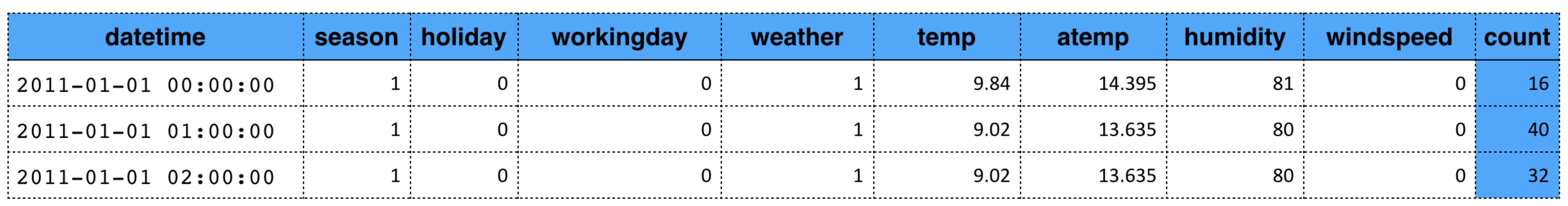}\label{a}}\\
\subfloat[Additionally engineered features using our technique.]{\includegraphics[width=\linewidth]{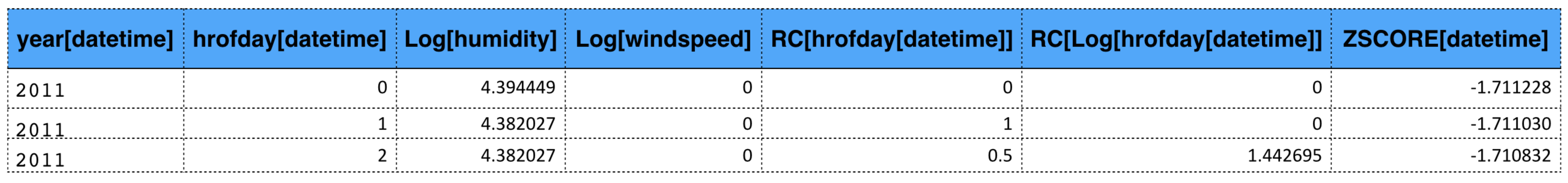}\label{b}}\\
\caption{In Kaggle's biking rental count prediction dataset, FE through our technique reduced Relative Absolute Error from 0.61 to 0.20 while retaining interpretability of features.}
\label{fig:bikingexample}
\end{figure}

For instance, Figure~\ref{fig:ex1} depicts two different representations for points belonging to a classification problem dataset. On the left, one can see that instances corresponding to the two classes are present in alternating small clusters. For most machine learning (ML) algorithms, it is hard to draw a reasonable classifier on this representation that separates the two classes On the other hand if the feature $x$ is replaced by its $sine$, as seen in the image on the right, it makes the two classes reasonably separable by most classifiers. 
The {\em task} or {\em process} of altering the feature representation of a predictive modeling problem to better fit a training algorithm is called {feature engineering} (FE). The {\em sine} function is an instance of a {\em transformation} used to perform FE.
Consider the schema of a dataset for forecasting hourly bike rental demand\footnote{\url{https://www.kaggle.com/c/bike-sharing-demand}} in Figure~\ref{fig:bikingexample}(a).
Deriving several features (Figure~\ref{fig:bikingexample}(b)) dramatically reduces the modeling error. For instance, extracting the hour of the day from the given timestamp feature helps to capture certain trends such as peak versus non-peak demand. Note that some valuable features are derived through a composition of multiple simpler functions. FE is perhaps the central task in improving predictive modeling performance, as documented in a detailed account of the top performers at various Kaggle competitions~\cite{Wind14}. 


%

In practice, FE is orchestrated by a data scientist, using hunch, intuition and domain knowledge based on continuously observing and reacting to the model performance through {\em trial and error}. As a result, FE is often time-consuming, and is prone to bias and error. Due to this inherent dependence on human decision making, FE is colloquially referred to as ``{\em an art/science}''~\footnote{\url{http://www.datasciencecentral.com/profiles/blogs/feature-engineering-tips-for-data-scientists}}~\footnote{\url{https://codesachin.wordpress.com/2016/06/25/non-mathematical-feature-engineering-techniques-for-data-science/}}, making it difficult to automate. The existing approaches to automate FE are either computationally expensive evaluation-centric and/or lack the capability to discover complex features.

We present a novel approach to automate FE based on reinforcement learning (RL). It involves training an agent on FE examples to learn an effective strategy of exploring available FE choices under a given budget. The learning and application of the exploration strategy is performed on a {\em transformation graph}, a directed acyclic graph representing relationships between different transformed versions of the data.
To the best of our knowledge, this is the first work that learns a performance-guided strategy for effective feature transformation from historical instances. Also, this is the only work in FE space that provides an adaptive budget constrained solution. Finally, the output features are compositions of well-defined mathematical functions which make them human readable and usable as insights into a predictive analytics problem, such as the one illustrated in Figure~\ref{fig:bikingexample}(b). 

\section{Related Work}
\label{sec:relatedwork}
Given a supervised learning dataset, FICUS~\cite{MarkovitchRosenstein02} performs a beam search over the space of possible features, constructing new features by applying ``constructor functions" (e.g. inserting an original feature into a composition of transformations). FICUS's search for better features is guided by heuristic measures based on information gain in a decision tree, and other surrogate measures of performance. In constrast, our approach optimizes for the prediction performance criterion directly, rather than surrogate criteria, and that we require no constructor functions. 
Note that FICUS is more general than a number of less recent approaches~\cite{RagRenShaTes93,BagalloHaussler90,YanRenBli91,MatheusRendell89,HuKibler96}.

Fan et al.~\cite{FanZhoPen10} propose FCTree uses a decision tree to partition the data using both original and constructed features as splitting points (nodes in the tree). As in FICUS~\cite{MarkovitchRosenstein02}, FCTree uses surrogate tree-based information-theoretic criteria to guide the search, as opposed to the true prediction performance. FCTree is capable of generating only simple features, and is not capable of composing transformations, i.e. it is search in a smaller space than our approach. They also propose a weight update mechanism that helps identify good transformations for a dataset, such that they are used more frequently.

The Deep Feature Synthesis component of Data Science Machine (DSM)~\cite{kanter:dsm} relies on applying all transformations on all features at once (but no combinations of transformations), and then performing feature selection and model hyper-parameter optimization over the combined augmented dataset. 
A similar approach is adopted by One Button Machine~\cite{lam2017one}. We will call this category as the {\em expansion-reduction} approach.
This approach suffers performance performance and scalability bottleneck due to performing feature selection on a large number of features that are explicitly generated by simultaneous application of all transforms. In spite of the expansion of the explicit expansion of the feature space, it does not consider the composition of transformations.

FEADIS~\cite{DorReich12} relies on a combination of random feature generation and feature selection. 
It adds constructed features greedily, and as such requires many expensive performance evaluations.
A related work, ExploreKit~\cite{DBLP:conf/icdm/KatzSS16} expands the feature space explicitly. It employs learning to rank the newly constructed features and evaluating the most promising ones. While this approach is more scalable than the expand-select type, it still is limited due to the explicit expansion of the feature space, and hence time-consuming. For instance, their reported results were obtained after running FE for days on moderately sized datasets. Due to the complex nature of this method, it does not consider compositions of transformations. We refer to this approach of FE as {\em evolution-centric}. 

Cognito~\cite{khurana2016cognito} introduces the notion of a tree-like exploration of transform space; they present a few simple handcrafted heuristics traversal strategies such as breadth-first and depth-first search that do not capture several factors such as adapting to budget constraints. This paper generalizes the concepts introduced there.
LFE~\cite{lfe} proposes a learning based method to predict the most likely useful transformation for each feature. It considers features independent of each other; it is demonstrated to work only for classification so far, and does not allow for composition of transformations.

Other plausible approaches to FE are hyper-parameter optimization~\cite{hyperopt} where each transformation choice could be a parameter, black-box optimization strategies~\cite{li2016efficient}, or bayesian optimization such as the ones for model- and feature-selection~\cite{feurer:autosklearn}. To the best of our knowledge, these approaches have been employed for solving FE. ~\cite{Smith2003} employ a genetic algorithm to determine a suitable transformation for a given data set, but is limited to single transformations.

Certain ML methods perform some level of FE indirectly.
A recent survey on the topic appears can be found here~\cite{StoRosKu15}.
Dimensionality reduction methods such as Principal Component Analysis (PCA) and its non-linear variants (Kernel PCA)~\cite{Fodor02} aim at mapping the input dataset into a lower-dimensional space with fewer features.
Such methods are also known as \textit{embedding} methods~\cite{StoRosKu15}. 
Kernel methods~\cite{ShaweTaylorCristianini04} such as Support Vector Machines (SVM) are a class of learning algorithms that use kernel functions to implicitly map the input feature space into a higher-dimensional space. 

Multi-layer neural networks allow for useful features to be learned automatically, such that they minimize the training loss function. Deep learning methods have made remarkable successes on various data such as video, image and speech, where manual FE is very tedious.~\cite{Bengio13}. 
However, deep learning methods require massive amounts of data to avoid overfitting and are not suitable for problems instances of small or medium sizes, which are quite common. 
Additionally, deep learning has mostly been successful with video, image, speech and natural language data, whereas the general numerical types of data encompasses a wide variety of domains and need FE. Our technique is domain, and model independent and works generally irrespective of the scale of data.
Also, the features learned by a deep network may not always be easily explained, limiting application domains such as healthcare~\cite{ChePurKheLiu15}. On the contrary, features generated by our algorithm are compositions of well-understood mathematical functions that can be analyzed by a domain expert. 



\section{Overview}
\label{sec:overview}
The automation of FE is challenging computationally, as well as in terms of decision-making. First, the number of possible features that can be constructed is unbounded since transformations can be composed, i.e., applied repeatedly to features generated by previous transformations. In order to confirm whether a new feature provides value, it requires training and validation of a new model upon including the feature. It is an expensive step and infeasible to perform with respect to each newly constructed feature. The  {\em evolution-centric} approaches described in the Related Work section operate in such manner and take days to complete even on moderately-sized datasets. Unfortunately, there is no reusability of results from one evaluation trial to another. On the other hand, the {\em expansion-reduction} approach performs fewer or only one training-validation attempts by first explicitly applying all transformations, followed by feature selection on the large pool of features. It presents a scalability and speed bottleneck itself; in practice, it restricts the number of new features than can be considered. In both cases, there is a lack of performance oriented search. With these insights, our proposed framework performs a systematic enumeration of the space of FE choices for any given dataset through a {\em transformation graph}. Its nodes represent different versions of a given datasets obtained by the application of transformation functions (edges). A transformation when applied to a dataset, applies the function on all possible features (or sets of features in case non-unary functions), and produces multiple additional features, followed by optional feature selection, and training-evaluation. Therefore, it {\em batches} the creation of new features by each transformation function. This lies somewhat in the middle of the evolution-centric and the expansion-reduction approaches. It not only provides a computation advantage, but also a logical unit of measuring performance of various transforms, which is used in composing different functions in a performance-oriented manner. This translates the FE problem to finding the node (dataset) on the transformation graph with the highest cross-validation performance, while only exploring the graph as little as possible. 
It also allows for composition of transformation functions. 

Secondly, the decision making in manual FE exploration involves initiation and complex associations, that are based on a variety of factors. Some examples are: prioritizing transformations based on the performance with the given dataset or even based on past experience; whether to {\em explore} different transformations or {\em exploit} the combinations of the ones that have shown promise thus far on this dataset, and so on. It is hard to articulate the notions or set of rules that are the basis of such decisions; hence, we recognize the factors involved and {\em learn} a strategy as a function of those factors in order to perform the exploration automatically. 
We use reinforcement learning on FE examples on a variety of datasets, to find an optimal strategy. This is based on the transformation graph. The resultant strategy is a policy that maps each instance of the transformation graph to the action of applying a transformation on a particular node in the graph. 


\subsection{Notation and Problem Description}
Consider a predictive modeling task consisting of (1) a set of features, $F = \{f_1, f_2 \dots f_m\}$; (2) a target vector, $y$. A pair of the two is specified as a dataset, $D=\langle F,y \rangle$. The nature of $y$, whether categorical or continuous, describes if it is a classification or regression problem, respectively. For an applicable choice of learning algorithm $L$ (such as Random Forest or Linear Regression) and a measure of performance, $m$ (such as AUROC or -RMSE). We use $A_L^m(F,y)$ (or simply, $A(F)$ or $A(D)$) to signify the cross-validation performance of a the model constructed on given data with using the algorithm $L$ through the performance measure $m$.

Additionally, consider a set of $k$ transformation functions at our disposal, $\mathcal T = \{t_1, t_2 \dots t_k\}$. 
The application of a transformation on a set of features, $t(F)$, suggests the application of the corresponding function on all valid input feature subsets in $F$, applicable to $t$. 
For instance, a $square$ transformation applied to a set of features, $F$ with eight numerical and two categorical features will produce eight new output features, $f_o = square(f_i), \forall f_i \in F, f_i \in \mathbb{R}^n$. 
This extends to $k$-ary functions, which work on $k$ input features.
A derived feature is recognized with a hat, $\hat{f}$. 
The entire (open) set of derived features from $F$ is denoted by $\hat{F}$.

A `+' operator on two feature sets (associated with the same target $y$) is a union of the two feature sets, $F_o = F_1 + F_2 = F_1 \cup F_2$, preserving row order. 
Generally, transformations add features; on the other hand, a {\em feature selection} operator, which is to a transformation in the algebraic notation, removes features. 
Note that all operations specified on a feature set , $T(F)$, can exchangeably be written for a corresponding dataset, $D=\langle F,y \rangle$, as, $T(D)$, where it is implied that the operation is applied on the corresponding feature set. Also, for a binary, such as sum, $D_o = D_1+D_2$, it is implied that the target is common across the operands and the result.



\label{subsec:probdesc}
The goal of feature engineering is stated as follows. Given a set of features, $F$, and target, $y$, find a set of features, $F^{*}$ where each feature is in $F^{*}=F_1 \cup F_2$, where $F_1 \subseteq F$ (original) and $F_2 \subset \hat{F}$ (derived), to maximize the modeling accuracy for a given algorithm, $L$ and measure, $m$.

\begin{equation} \label{eq:problem}
F^* = \argmax_{F_1,F_2}{A_L^m(F_1 \cup F_2, y)}
\end{equation}

\vspace{-1pt}
\section{Transformation Graph}
\label{subsec:tg}
{\em Transformation Graph}, $G$, for a given dataset, $D_0$, and a finite set of transformations, $T$, is a directed acyclic graph in which each node corresponds to a either $D_0$ or a dataset derived from it through transformation path. Every node's dataset contains the same target and number of rows. The nodes are divided into three categories: (a) The start or the root node, $D_0$ corresponding to the given dataset; (b) Hierarchical nodes, $D_i$, where $i > 0$, which have one and only one incoming node from a parent node $D_j$, $i > j \ge 0$, and the connecting edge from $D_j$ to $D_i$ corresponds to a transform $T \in \mathcal T$ (including feature selection), i.e., $D_j = T (D_i)$; (c) Sum nodes, $D_{i,j}^+ = D_i + D_j$, a result of a dataset sum such that $i \ne j$. Similarly, edges correspond to either transforms or `+' operations, with children as type (b) or type (c) nodes, respectively. The direction of an edge represents the application of transform from  source to a target dataset (node). Height ($h$) of the transformation graph refers to the maximum distance between the root and any other node.

\begin{figure}
\includegraphics[width=0.4\textwidth]{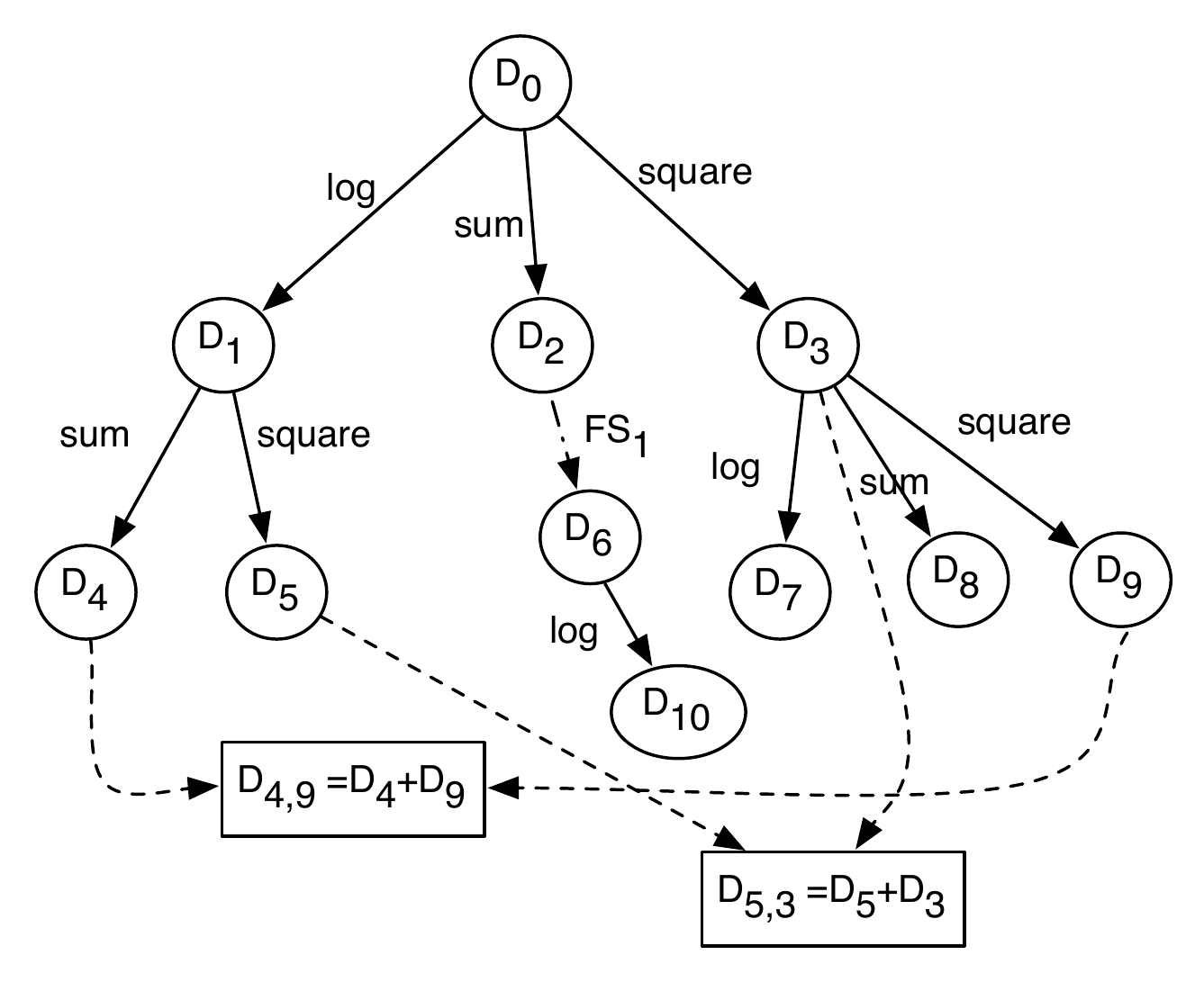}
\caption{Example of a Transformation Graph (a DAG). The start node $D_0$ corresponds to the given dataset; that and the hierarchical nodes are circular. The sum nodes are rectangular. In this example, we can see three transformations, {\em log}, {\em sum}, and {\em square}, as well as a feature selection operator $FS_1$.}
\label{fig:tgex}
\end{figure}

A transformation graph is illustrated in Figure~\ref{fig:tgex}. 
Each node of a transformation graph is a candidate solution for the FE problem in Equation~\ref{eq:problem}. Also, a complete transformation graph must contain a node that is the solution to the problem, through a certain combination of transforms including feature selection.
The operator $\theta (G)$ signifies all nodes of graph $G$, and $\theta_h(G)$ signifies all hierarchical nodes. Also, $\lambda (D_i,D_j)$ signifies the transformation $T$, such that its application on $D_i$ created $D_j$ as its child; alternatively if $D_j$ is a sum node and $D_i$ is one of its parents, then $\lambda (D_i,D_j) = +$.
A complete transformation graph is unbounded for a non-empty transformation set. A constrained (bounded height, $h$) complete transformation graph for $t$ transformations will have $t^{h+1}-2$ hierarchical nodes (and an equal number of corresponding edges), and $(t^{h+1}-1) \times (t^{h+1}-2) \over 2$ sum nodes (and 2 times corresponding edges). It can be seen that for even a height bounded tree with a modest number of transforms, computing the cross-validation performance across the entire graph is computationally expensive.

\subsection{Graph Exploration under a Budget Constraint}
\label{subsec:hgexplore}
It should be emphasized that the exhaustive exploration of a transformation graph is not an option, given its massive potential size. For instance, with 20 transformations and a height = 5, the complete graph contains about  3.2 million nodes; an exhaustive search would imply as many model training and testing iterations. On the other hand, there is no known property that allows us to deterministically verify the optimal solution in a subset of the trials. Hence, the focus of this work is to find a performance driven exploration policy, which maximizes the chances of improvement in accuracy within in a {\em limited time budget}. 
The exploration of the transformation graph begins with a single node, $D_0$, and grows one node at a time from the then current state of the graph. 
In the beginning it is reasonable to perform {\em exploration} of the environment, i.e., stumble upon the transforms that signal an improvement. Over time (elapsed budget) it is desirable to reduce the amount of exploration and focus more on {\em exploitation}.

\begin{algorithm}
{\bf Input:} Dataset $D_0$, Budget $B_{max}$\;
Initialize $G_0$ with root $D_0$\;
 \While{$i < B_{max}$}{
  $\mathcal N \gets \theta(G_i)$\; 
  $b_{ratio} = {i \over B_{max}}$
\begin{flalign*}
&n^*,t^* \gets \operatorname*{arg\,max}_{n,t \nexists n'  \forall t=\lambda(n,n')} R(G_i, n, t, b_{ratio})&
\end{flalign*}%

$G_{i+i} \gets$ Apply $t^*$ to $n^*$ in $G_{i}$
%
%

$i \gets i+1$\;
 }
{\bf Output:}  $\underset{D}{\textrm{argmax}}$ $A(\theta(G_i))$
 
 \caption{Transformation Graph Exploration}
 \label{algo1}
\end{algorithm}

Algorithm~\ref{algo1} outlines our general methodology for exploration. At each step, an estimated reward from each possible move, $R(G_i,n,t, {i \over B_{max}})$ is used to rank the options of actions available at each given state of the transformation graph $G_i, \forall i \in [0,B_{max})$, where $B_{max}$ is the overall allocated budget in number of steps\footnote{Budget can be considered in terms of any quantity that is monotonically increasing in $i$, such as time elapsed. For simplicity, we work with ``number of steps''.}. Note that the algorithm allows for different exploration strategies, which is left up to the definition of the function $R(\dots)$, which defines the relative importance of different steps at each step. The parameters of the function suggest that it depends on the various aspects of the graph at that point, $G_i$, the remaining budget, and specifically, attributes of the action (node+transformation) being characterized. Below, we briefly discuss such factors that influence the exploration choice at each step. These factors are compared across all choices of node-transformation pairs $<n,t>$ at $G_i$:


\begin{enumerate}[topsep=0pt,itemsep=-1ex,partopsep=1ex,parsep=1ex]
\item Node $n$'s Accuracy: Higher accuracy of a node incentives further exploration from that node, compared to others. 
\item Transformation, $t$'s average immediate reward till $G_i$.
\item Number of times transform $t$ has already been used in the path from root node to $n$.
\item Accuracy gain for node $n$ (from its parent) and gain for $n$'s parent, i.e., testing if $n$'s gains are recent.
\item Node Depth: A higher value is used to penalize the relative complexity of the transformation sequence.
\item The fraction of budget exhausted till $G_i$.
\item Ratio of feature counts in $n$ to the original dataset: This indicates the bloated factor of the dataset.
\item Is the transformation a feature selector? 
\item Whether the dataset contain (a) numerical features; or (b) datetime features; or (c) string features?
\end{enumerate}

Simple graph traversal strategies can be handcrafted. A strategy essentially translates to the design of the reward estimation function, $R(\dots)$. In line with Cognito~\cite{khurana2016cognito}, a breadth-first or a depth-first strategy, or perhaps a mix of them can be described. While such simplistic strategies work suitably in specific circumstances, it seems hard to handcraft a unified strategy that works well under various circumstances. We instead turn to machine learning to learn the complex strategy from several historical runs.

\section{Traversal Policy Learning}
\label{subsec:rl}

So far, we have discussed a hierarchical organization of FE choices through a transformation graph and a general algorithm to explore the graph in a budget allowance. At the heart of the algorithm is a function to estimate reward of an action at a state. The design of the reward estimation function determines the strategy of exploration. Strategies could be handcrafted; however, in this section we try to learn an optimal strategy from examples of FE on several datasets through transformation graph exploration. 
Because of the behavioral nature of this problem - which can be perceived as continuous decision making (which transforms to apply to which node) while interacting with an environment (the data, model, etc.) in discrete steps and observing reward (accuracy improvement), with the notion of a final optimization target (final improvement in accuracy), it is modeled as a RL problem. We are interested in learning an {\em action-utility} function to satisfy the expected reward function, R(\dots) in Algorithm~\ref{algo1}. In the absence of an explicit model of the environment, we employ Q-learning with function approximation due to the large number of states (recall, millions of nodes in a graph with small depth) for which it is infeasible to learn state-action transitions explicitly. 

Consider the graph exploration process as a {\em Markov Decision Process (MDP)} where the {\em state} at step $i$ is a combination of two components: (a) transformation graph after $i$ node additions, $G_i$ ($G_0$ consists of the root node corresponding to the given dataset. $G_i$ contains $i$ nodes); (b) the remaining budge at step $i$, i.e., $b_{ratio} = {i \over B_{max}}$. 
Let the entire set of states be $S$. On the other hand, 
an {\em action} at step $i$ is a pair of existing tree node and transformation, i.e., $<n,t>$ where $n \in \theta(G_t)$, $t \in T$ and $\nexists n \in G_i \forall \lambda(n, n') = t$; it signifies the application of the one transform (which hasn't already been applied) to one of the exiting nodes in the graph. Let the entire set of actions be $C$. A policy, $\Pi : S \rightarrow C$, determines which action is taken given a state. Note that the objective of RL here is to learn the optimal policy (exploration strategy) by learning the action-value function, which we elaborate later in the section.

Such formulation uniquely identifies each state. Considering the ``remaining budget'' as factor in the state of the MDP helps us address the {\em runtime exploration versus exploitation} trade-off for a given dataset. 
Note that this runtime explore/exploit trade-off is not identical to the commonly referred trade-off in RL training in context of selecting actions to balance reward and not getting stuck in a local optimum.

At step $i$, the occurrence of an action results in a new node, $n_i$, and hence a new dataset on which a model is trained and tested, and its accuracy $A(n_i)$ is obtained. 
To each step, we attribute an immediate scalar reward:
\[r_i = \max_{n' \in \theta(G_{i+1})}{A(n')} - \max_{n \in \theta(G_i)}{A(n)}\]
with $r_0=0$, by definition.
 The cumulative reward over time from state $s_i$ onwards is defined as: 
 \[R(s_i) = \sum_{j=0}^{B_{max}}{\gamma^i. r_{i+j}}\]
where $\gamma \in [0,1)$ is a discount factor, which prioritizes early rewards over the later ones. The goal of RL here is to find the optimal policy $\Pi^*$ that maximizes the cumulative reward.

We use Q-learning~\cite{watkins1992q} with function approximation to learn the action-value Q-function. For each state, $s \in S$ and action, $c \in C$, Q-function with respect to policy $\Pi$ is defined as:
\[Q(s, c) = r(s, c) + \gamma R^\Pi {(\delta(s,c))} \]
where $\delta : S \times C \rightarrow S$ is a hypothetical transition function, and $R^\Pi(s)$ is the cumulative reward following  state $s$. The optimal policy is:
\begin{equation} \label{eq:optpol}
\Pi^*(s) = \operatorname*{arg\,max}_{c}[Q(s,c)]
\end{equation}

However, given the size of $S$, it is infeasible to learn Q-function directly. Instead, a linear approximation the Q-function is used as follows:
\begin{equation} \label{eq:approx}
Q(s,c) =  w^c . f(s)
\end{equation}

 where $w^c$ is a weight vector for action $c$ and $f(s) = f(g,n,t,b)$ is a vector of the state characteristics described in the previous section and the remaining budget ratio. Therefore, we approximate the Q-functions with linear combinations of characteristics of a state of the MDP. 
Note that, in each heuristic rule strategy, we used a subset of these state characteristics, in a self-conceived manner. However, in the ML based approach here, we select the entire set of characteristics and let the RL find the appropriate weights of those characteristics (for different actions). Hence, this approach generalizes the other handcrafted approaches.

The update rule for $w_c$ is as follows:
\begin{equation} \label{eq:1}
w^{c_j} \gets w^{c_j}+ \alpha . (r_j + \gamma . \max_{n',t'} Q(g',c') - Q(g,c)). f(g,b) 
\end{equation}
where $g'$ is the state of the graph at step $j+1$, and $\alpha$ is the learning rate parameter. The proof follows from~\cite{irodova2005reinforcement}.

A variation of the linear approximation where the coefficient vector $w$ is independent of the action $c$, is as follows:
\begin{equation} \label{eq:approx2}
Q(s,c) =  w . f(s)
\end{equation}

This method reduces the space of coefficients to be learnt by a factor of $c$, and makes it faster to learn the weights. It is important to note that the Q-function is still not independent of the action $c$, as one of the factors in $f(s)$ or $f(g,n,t,b)$ is actually the average immediate reward for the transform for the present dataset. Hence, Equation~\ref{eq:approx2} based approximation still distinguishes between various actions ($t$) based on their performance in the transformation graph exploration so far; however, it does not learn a bias for different transformations in general and based on the feature types (factor \#9). We refer to this type of strategy as $RL_2$. In our experiments RL2 efficiency is somewhat inferior to the strategy to the strategy learned with Equation~\ref{eq:approx}, which we refer to as $RL_1$.

\begin{table*}[t]
\begin{center}
  \begin{tabular}{| l||  r  r  r  r || r  r   r  r  r  |}
    \hline
    Dataset & Source & C/R  & Rows & Features & Base & Our $RL_1$  & Exp-Red & Random & Tree-Heur    \\ \hline     \hline
 Higgs Boson & UCIrvine & C & 50000& 28 & 0.718 &  \bf{0.729}  &  0.682 & 0.699  & 0.702     \\ \hline
      Amazon Employee & Kaggle & C & 32769 & 9 & 0.712 & \bf{0.806} & 0.744& 0.740   & 0.714   \\ \hline
   PimaIndian  & UCIrvine & C & 768 & 8 &  0.721 & \bf{0.756}  & 0.712 & 0.709  & 0.732   \\ \hline
  SpectF        & UCIrvine & C & 267 & 44 &  0.686 & {0.788}  & \bf0.790 & 0.748  & 0.780  \\ \hline
    SVMGuide3 & LibSVM & C & 1243 & 21 & 0.740 & \bf{0.776} &0.711 & 0.753  & \bf{0.766}    \\ \hline
     German Credit & UCIrvine & C & 1001 & 24 &0.661 & \bf{0.724} &  0.680 &0.655  & 0.662   \\ \hline
      Bikeshare DC & Kaggle & R & 10886 & 11 & 0.393 &\bf{0.798} &  0.693 & 0.381 & 0.790         \\ \hline
      Housing Boston & UCIrvine & R   & 506 & 13 & 0.626 & \bf{0.680} & 0.621 &0.637& 0.652           \\ \hline
      Airfoil  & UCIrvine   &R&  1503 & 5 & 0.752 & {\bf 0.801} & 0.759 &0.752 & 0.794               \\ \hline
      AP-omentum-ovary & OpenML & C & 275 & 10936 & 0.615 &  \bf{0.820} &  0.725& 0.710  & 0.758    \\ \hline
      Lymphography & UCIrvine & C & 148 & 18 & 0.832 & \bf{0.895} & 0.727 & 0.680   & 0.849 \\ \hline
       Ionosphere & UCIrvine & C & 351& 34 & 0.927 & \bf{0.941} &  {0.939} & 0.934   & \bf{0.941} \\ \hline
       Openml\_618 & OpenML  &R  & 1000 & 50 & 0.428 &  \bf{0.587} & 0.411   & 0.428   & 0.532 \\ \hline
        Openml\_589 & OpenML  &R   & 1000 & 25 & 0.542 &  \bf{0.689}  & 0.650 & 0.571   & 0.644      \\ \hline
    Openml\_616 & OpenML   &R & 500 & 50 & 0.343 &  \bf{0.559}  & 0.450    & 0.343     & 0.450       \\ \hline    
    Openml\_607 & OpenML   &R & 1000 & 50 & 0.380 &  \bf{0.647}    & 0.590   & 0.411    & 0.629         \\ \hline     
    Openml\_620 & OpenML   &R & 1000 & 25 & 0.524 &  \bf{0.683}    & 0.533   & 0.524    & 0.583          \\ \hline     
     Openml\_637 & OpenML   &R & 500 & 50 & 0.313 &  \bf{0.585}    & 0.581    & 0.313    & 0.582       \\ \hline     
     Openml\_586 & OpenML   &R & 1000 & 25 & 0.547 &  \bf{0.704}    & 0.598    & 0.549    & 0.647       \\ \hline     
         Credit Default & UCIrvine & C & 30000& 25 & 0.797 & \bf{0.831}   & 0.802   & 0.766    & 0.799        \\ \hline     
        Messidor\_features & UCIrvine & C & 1150& 19 & 0.691 &  \bf{0.752}    & 0.703    & 0.655  &  0.762      \\ \hline     
         Wine Quality Red & UCIrvine & C & 999& 12 & 0.380 &  \bf{0.387}    & 0.344   &0.380 &  0.386      \\ \hline     
         Wine Quality White & UCIrvine & C & 4900& 12 & 0.677 & \bf{0.722}   & 0.654   & 0.678  & 0.704       \\ \hline     
            SpamBase & UCIrvine & C & 4601& 57 & 0.955 & \bf{0.961}   & 0.951  &  0.937  & {0.959}            \\ \hline     
  \end{tabular}
  \caption{Comparing performance for base dataset (no FE), Our FE, Expansion-Reduction style FE, Random FE, and Tree heuristic FE, using 24 datasets. Performance here is FScore for classification (C) and  ($1-$rel. absolute error) for regression (R). }
  \label{tab:expt1}
\end{center}
\end{table*}

\section{Experiments}
\label{sec:experiments}
\topic{Training:} We used 48 datasets (not overlapping with test datasets) to select training examples 
using different values for maximum budget, $B_{max} \in \{25,50,75,100,150,200,300,500\}$ with each dataset, in a random order. We used the discount factor, $\gamma = 0.99$, and learning rate parameter, $\alpha =0.05$. The weight vectors, $w^c$ or $w$, each of size 12, were initialized with 1's. The training example steps are drawn randomly with the probability $\epsilon=0.15$ and the current policy with probability $1-\epsilon$.
We have used the following transformation functions in general (except when specified a smaller number): {\em Log, Square, Square Root, Product, ZScore, Min-Max-Normalization, TimeBinning, Aggregation (using Min,Max,Mean,Count,Std), Temporal window aggregate, Spatial Aggregation, Spatio Temporal Aggregation, k-term frequency, Sum, Difference, Division, Sigmoid, BinningU, BinningD, NominalExpansion, Sin, Cos, TanH}.

\topic{Comparison:}
We tested the impact of our FE on 24 publicly available datasets (different from the datasets used in training) from a variety of domains, and of various sizes.  We report the accuracy of (a) {\em base} dataset; (b) {\em Our FE} routine with $RL_1$, $B_{max}=100$; (c) {\em Expansion-reduction} implementation where all transformations are first applied separately and add to original columns, followed by a feature selection routine; 
(d) {\em Random:} randomly applying a transform function to a random feature(s) and adding the result to the original dataset and measuring the CV performance; this is repeated 100 times and finally, we consider all the new features whose cases showed an improvement in performance, along with the original features to train a model (e) {\em Tree-Heur}: our implementation of Cognito's~\cite{khurana2016cognito} global search heuristic for $100$ nodes. We used {\em Random Forest} with default Weka parameters as our learning algorithm for all the comparisons as it gave us the strongest baseline (no FE) average. A 5-fold cross validation using random stratified sampling was used. The results for a representative are captured in Table~\ref{tab:expt1}. It can be seen that our FE outperforms others in most of the cases but one (where expand-reduce is better) and tied for two with Cognito global search.
Our technique reduces the error (relative abs. error, or 1- mean unweighted FScore) by $23.8\%$ (by median) for the 24 datasets presented in Table~\ref{tab:expt1}.  
For reference to time taken, it took the Bikeshare DC dataset 4 minutes, 40 seconds to run for 100 nodes for our FE, on a single thread on a 2.8GHz processor. Times taken by the Random and Cognito were similar to our FE for all datasets, while expand-reduce took 0.1 to 0.9 times the time of our FE, for different datasets.

\begin{figure}
\includegraphics[width=0.45\textwidth]{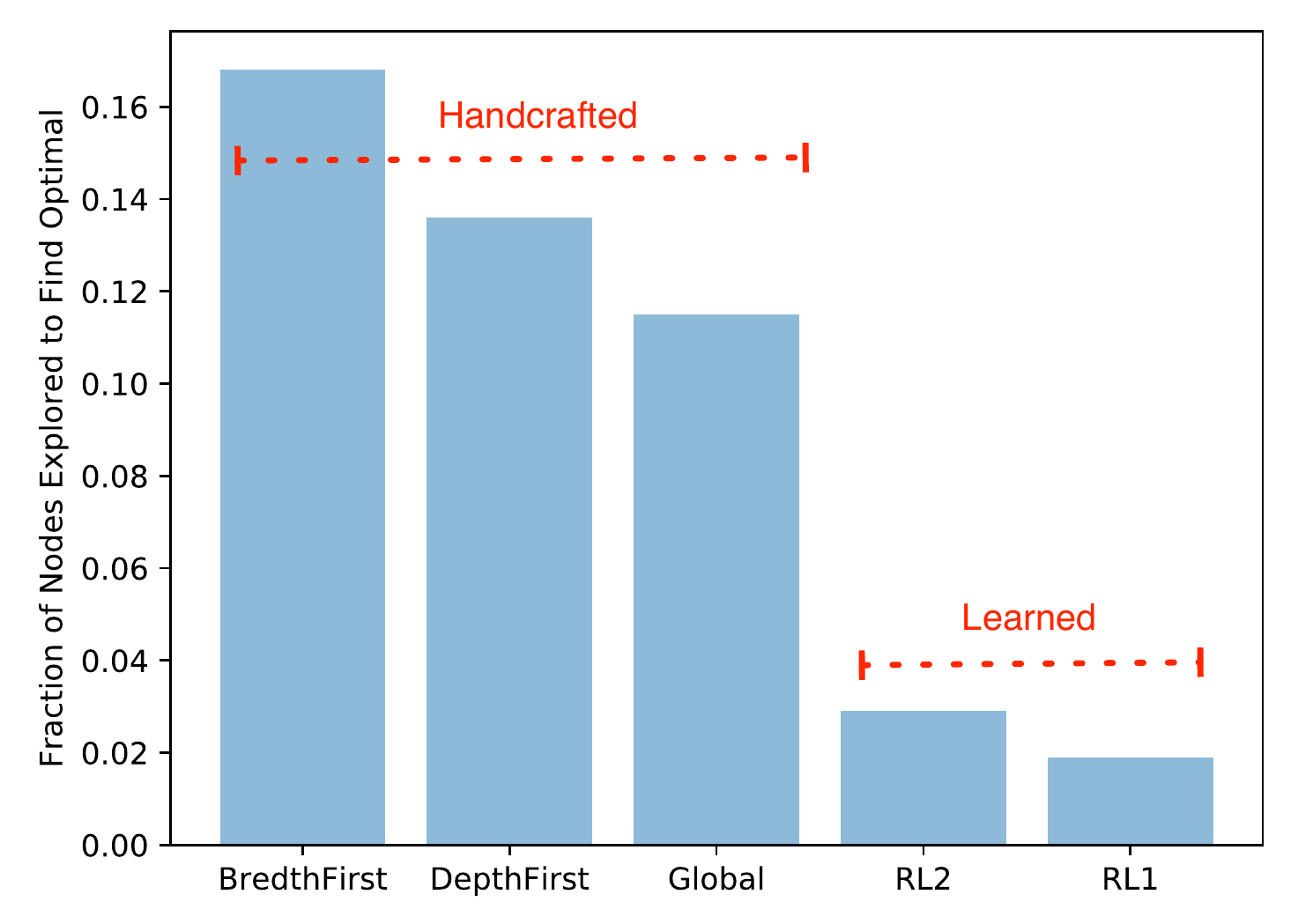}
\caption{Comparing efficiencies of exploration policies.}
\label{fig:policyexpt}
\end{figure}

\begin{figure}[h]
\includegraphics[width=0.45\textwidth]{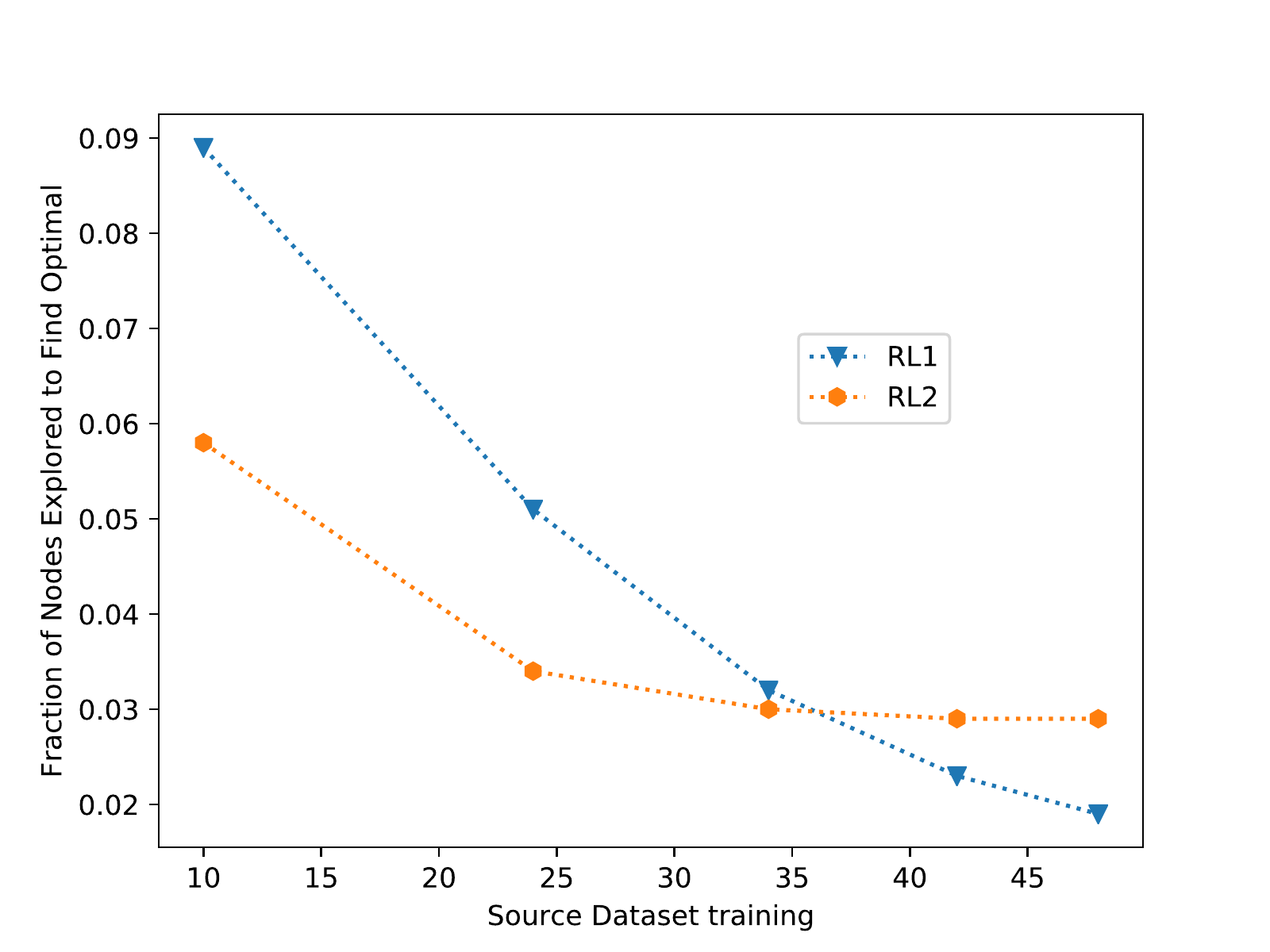}
\caption{Policy effectiveness with training dataset sources}
\label{fig:training}
\end{figure}

\topic{Traversal Policy Comparison:} In Figure~\ref{fig:policyexpt}, we see that on an average for 10 datasets, the RL-based strategies are 4-8 times more efficient than any handcrafted strategy ({\em breadth-first, depth-first} and {\em global} as described in~\cite{khurana2016cognito}), in finding the optimal dataset in a given graph with 6 transformations and bounded height, $h_{max}=4$. Also, Figure~\ref{fig:training} tells us that while $RL_1$ (Eqn.~\ref{eq:approx}) takes more data to train, it is more efficient than $RL_2$ (Eqn.~\ref{eq:approx2}), demonstrating that learning a general bias for transformations and one conditioned on data types makes the exploration more efficient.

\begin{figure}[h]
\includegraphics[width=0.45\textwidth]{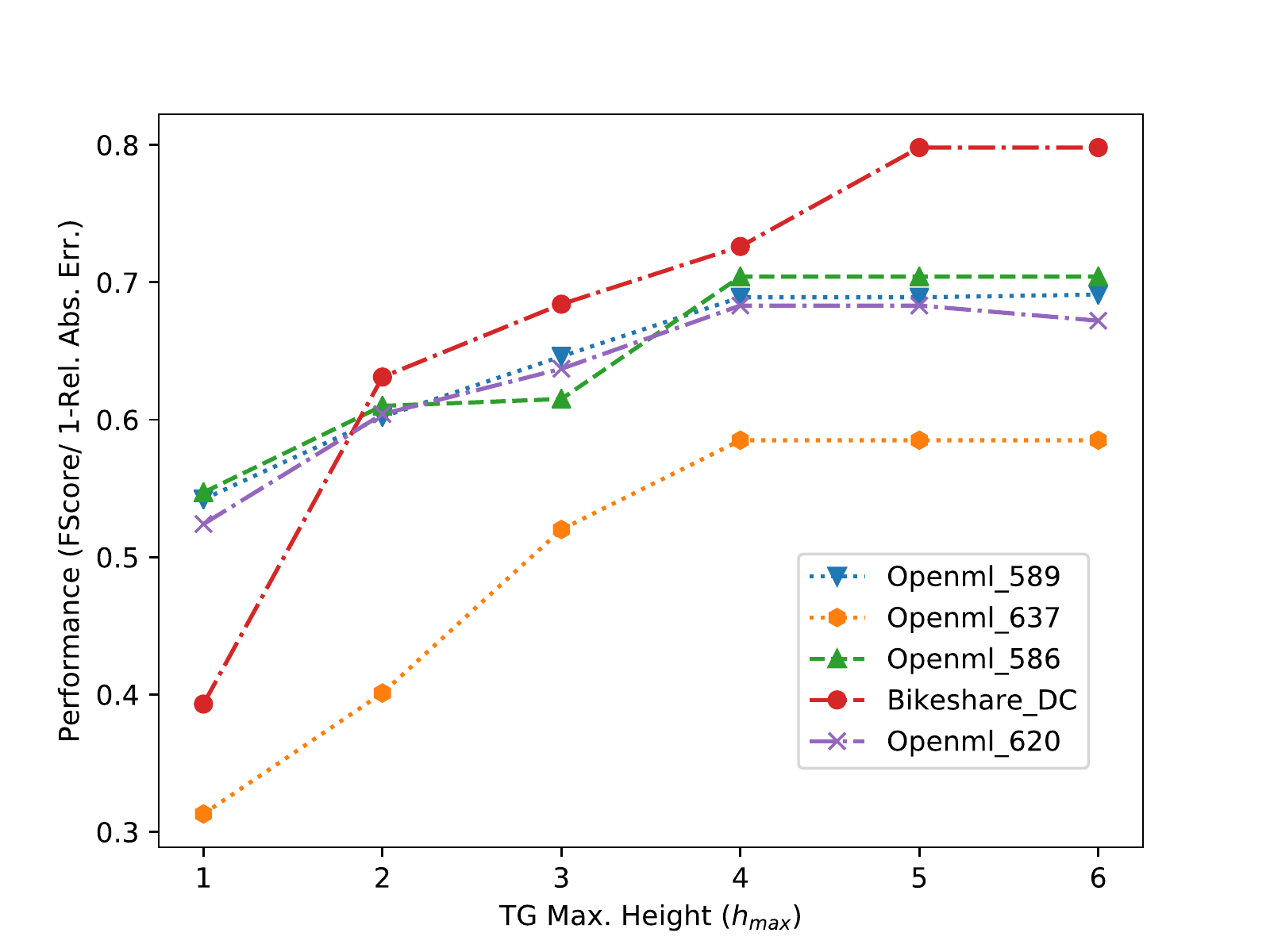}
\caption{Performance vs. $h_{max}$}
\label{fig:expH}
\end{figure}

\topic{Internal System Comparisons:} We additionally performed experimentation to test and tune the internals of our system. Figure~\ref{fig:expH} shows the maximum accuracy node (for 5 representative datasets) found when the height was constrained to a different numbers, using $B_{max}=100$ nodes; $h_{max}=1$ signifies base dataset. Majority of datasets find the maxima with $h_{max}=4$ with most find it with $h_{max}=5$. For $h_{max}=6$, a tiny fraction shows deterioration, which can be interpreted as unsuccessful exploration cost due to a higher depth. 
Also, using feature selection (compared to none) as a transform improves the final gain in performance by about 51\%, measured on the 48 datasets (aforementioned 24 + another 24).
Finally, the use of different models (learning algorithms) lead to different optimal features being engineered for the same dataset, even for similar improvements in performance.





\section{Conclusion and Future Work}
\label{sec:conclusion}
In this paper, we presented a novel technique to efficiently perform feature engineering for supervised learning problems. 
The cornerstone of our framework are, a transformation graph that enumerates the space of feature options and a RL-based, performance-driven exploration of the available choices to find valuable features. 
The models produced using our proposed technique considerably reduce the error rate (24\% by median) across a variety of datasets, for a relatively small computational budget. 
This methodology can potentially save a data analyst hours to weeks worth of time.
One direction to further improve the efficiency of the system is through a complex non-linear modeling of state variables. Additionally, extending the described framework to other aspects of predictive modeling, such as missing value imputation or model selection, is of potential interest as well. Since optimal features depend on model type (learning algorithm), a joint optimization of the two is particularly interesting.


\clearpage
\bibliographystyle{aaai}
\bibliography{cognito}

\end{document}